# Data-driven Sequential Monte Carlo in Probabilistic Programming

**Yura Perov**   **Tuan Anh Le**   **Frank Wood**
Department of Engineering Science, University of Oxford
{perov,tuananh,fwood}@robots.ox.ac.uk

## Abstract

Most of Markov Chain Monte Carlo (MCMC) and sequential Monte Carlo (SMC) algorithms in existing probabilistic programming systems suboptimally use only model priors as proposal distributions. In this work, we describe an approach for training a discriminative model, namely a neural network, in order to approximate the optimal proposal by using posterior estimates from previous runs of inference. We show an example that incorporates a data-driven proposal for use in a non-parametric model in the Anglican probabilistic programming system [9]. Our results show that data-driven proposals can significantly improve inference performance so that considerably fewer particles are necessary to perform a good posterior estimation.

## 1 Background

We consider a generative model $p(x_{1:N}, y_{1:N})$ with hidden variables $x_{1:N}$ and observations $y_{1:N}$. In probabilistic programming, we let the observing random variable $y_n$ be the value of the $n$th observe, and the hidden variables $\mathbf{x}_n = x_{1:n}$ be the execution trace before this observe. The goal of SMC inference in probabilistic programming is to sample from a family of distributions $p(\mathbf{x}_{1:n}|y_{1:n})$ for $n = 1, \ldots, N$. This is achieved by generating a set of particles $\{\mathbf{x}_{1:n}^{(p)}\}_{p=1}^P$ and the corresponding importance weights $\{w_n^{(p)}\}_{p=1}^P$. We approximate the target distribution as $\sum_{p=1}^P w_n^{(p)} \delta_{\mathbf{x}_{1:n}^{(p)}}(\mathbf{x}_{1:n})$. Particles at time $n$ are generated using a chosen proposal distribution $q_n(\mathbf{x}_n|\mathbf{x}_{n-1})$, which is used to propose those particles given the set of particles $\{\bar{\mathbf{x}}_{1:(n-1)}^{(p)}\}$ at the previous step $n-1$ that have been resampled from an SMC estimate of $p(\mathbf{x}_{1:(n-1)}|y_{1:(n-1)})$. The weights corresponding to these particles are then calculated as follows:

$$W_n^{(p)} = \frac{p\left(\mathbf{x}_n^{(p)}\big|\bar{\mathbf{x}}_{n-1}^{(p)}\right) p\left(y_n\big|\mathbf{x}_n^{(p)}\right)}{q\left(\mathbf{x}_n^{(p)}\big|\bar{\mathbf{x}}_{n-1}^{(p)}\right)}, \qquad w_n^{(p)} = \frac{W_n^{(p)}}{\sum_{i=1}^P W_n^{(i)}}, \qquad p = 1, \ldots, P.$$

Then, the proposal, resampling, and re-weighting steps are iterated for the next $n$.

In many applications of SMC inference, including probabilistic programming systems, the proposal distribution is taken to be the prior distribution $p(\mathbf{x}_n|\mathbf{x}_{n-1})$ of the generative model. This simplifies the implementation of such systems as we can readily sample and evaluate the densities of the proposed values directly from the generative model.

One of the problems SMC methods suffer from is high variance of weights and the related problem of propagation of low weight particles that waste computation. This is because the propose-from-prior method usually gives proposals far from the optimal one since the prior is significantly different from the posterior. These problems can be mitigated by resampling, which intuitively resets the system at the expense of the increase in the immediate Monte Carlo variance. However, resampling can introduce another problem of degenerate particle trajectories, in which only a few distinct values are used to represent the early part of the inferred trajectories. Another approach to minimise the variance of the weights is to use proposal distributions $\hat{q}_n(\mathbf{x}_n|\mathbf{x}_{n-1}, y_{1:n})$ that approximate the filtering distribution $p(\mathbf{x}_n|\mathbf{x}_{n-1}, y_{1:n})$ [2], from which it is assumed impossible to sample directly because the normalisation factor requires a marginalisation over $\mathbf{x}_n$. In particular, there are no known ways to sample from these distributions directly in probabilistic programming systems. In this paper, we explore using information from previous runs to improve the proposal distribution so that it approximates the distribution $p(\mathbf{x}_n|\mathbf{x}_{n-1}, y_{1:n})$.

## 2 Data-driven proposals for Sequential Monte Carlo

We want to have data-driven proposals for a certain subset of random choices $\mathcal{S} \subseteq \{1, \ldots, N\}$. We assume that the model prior $p(\mathbf{x}_n|\mathbf{x}_{n-1})$ has the same structure for all $n \in \mathcal{S}$ so that we can learn the same proposals for all of them. Although it is impossible to sample from the optimal proposal—the filtering distribution $p(\mathbf{x}_n|\rho_n)$ where the environment $\rho_n := (\mathbf{x}_{n-1}, y_{1:n})$—we can approximate it with a distribution $\hat{q}(\mathbf{x}_n|\mathfrak{N}_\theta(\phi(\rho_n)))$. This proposal distribution is parametrised by the output of some discriminative model $\mathfrak{N}_\theta$, which is in turn parametrised by $\theta$. The input of this model, namely features $\phi(\rho_n)$, is extracted from the environment $\rho_n$ by a fixed feature extractor function $\phi$.

To train the discriminative model, we need $M$ training inputs $\{\phi(\rho_n^{(i)})\}_{i=1}^{M}$ and related outputs $\{(\mathbf{x}_n^{(i)}, w_n^{(i)})\}_{i=1}^{M}$ such that each $\mathbf{x}_n^{(i)}$ is approximately drawn from the desired distribution $p\left(\mathbf{x}_n \middle| \rho_n^{(i)}\right)$. These weighted training inputs and outputs might be received from extensive sequential Monte Carlo inference on the training dataset.

The loss function used to obtain the parameters $\theta$ of the discriminative model is based on the Kullback-Leibler (KL) divergence. Taking the expectation of the KL-divergence with respect to $\mathbf{x}_{n-1}|y_{1:n}$, we can simplify the loss function $L_n(\theta)$ as follows:

$$L_n(\theta) = \mathbb{E}_{p(\mathbf{x}_{n-1}|y_{1:n})} \left[ D_{KL} \left( p(\mathbf{x}_n|\rho_n) \parallel \hat{q}(\mathbf{x}_n|\mathfrak{N}_\theta(\phi(\rho_n))) \right) \right]$$

$$= \mathbb{E}_{p(\mathbf{x}_{n-1}|y_{1:n})} \left[ \mathbb{E}_{p(\mathbf{x}_n|\mathbf{x}_{n-1}, y_{1:n})} \left[ \log \frac{p(\mathbf{x}_n|\rho_n)}{\hat{q}(\mathbf{x}_n|\mathfrak{N}_\theta(\phi(\rho_n)))} \right] \right]$$

$$= \mathbb{E}_{p(\mathbf{x}_n|y_{1:n})} \left[ \log \frac{p(\mathbf{x}_n|\rho_n)}{\hat{q}(\mathbf{x}_n|\mathfrak{N}_\theta(\phi(\rho_n)))} \right] = -\mathbb{E}_{p(\mathbf{x}_n|y_{1:n})} \left[ \log \hat{q}(\mathbf{x}_n|\mathfrak{N}_\theta(\phi(\rho_n))) \right] + c.$$

By ignoring the constant $c$ and substituting the Monte Carlo approximation of $p(\mathbf{x}_n|y_{1:n})$, $\sum_{i=1}^{M} w_n^{(i)} \delta_{\mathbf{x}_n^{(i)}}(\mathbf{x}_n)$, we get that

$$L_n(\theta) \approx -\sum_{i=1}^{M} w_n^{(i)} \log \hat{q}\left(\mathbf{x}_n^{(i)} \middle| \mathfrak{N}_\theta(\phi(\rho_n))\right). \tag{1}$$

The loss function has a convenient form for neural networks because it can be decomposed to a sum of losses corresponding to each neural network output. We also note that in practice, we might use training outputs $\{\mathbf{x}_n^{(i)}\}_{i=1}^{M}$ not from the filtering distribution $p\left(\mathbf{x}_n \middle| \mathbf{x}_{n-1}^{(i)}, y_{1:n}^{(i)}\right)$, but from the smoothing distribution $p\left(\mathbf{x}_n \middle| \mathbf{x}_{n-1}^{(i)}, y_{1:N}^{(i)}\right)$. This is because most of statistical inference in existing probabilistic programming systems is directed towards the approximation of the smoothing distributions.

## 3 Experiments

### 3.1 Dependent Dirichlet process mixture of objects

For further experiments, we have chosen the dependent Dirichlet process mixture of objects (DDPMO) model [7] in order to demonstrate applicability of our approach to more complicated models. The DDPMO is a recent Bayesian non-parametric model for detection-free tracking and object modelling. The DDPMO models the position and colour $\mathbf{x}_{t,n}$ of a foreground pixel $n$ at a video frame $t$ as an observed variable. This observed variable $\mathbf{x}_{t,n}$ depends on the latent variables of the model such as cluster assignments $c_{t,1:N_t}$ and object parameters $\theta_t^k$ for each cluster $k$. The DDPMO is a native Bayesian non-parametric model since the number of clusters and the related object parameters is unbounded and dependent on the observed data. The generative process of the DDPMO is described in the appendix.

The DDPMO model was implemented as an Anglican program. The full model source code has 120 lines, with comments. In addition, the generalised Pólya urn (GPU) procedure was written as an Anglican program. Its source code has 70 lines, with comments, and can be used for any other model in the future. In order to make inference tractable, we implement and employ exchangeable random procedures (XRP) for conjugate priors, which are the essential part of the generative process in the DDPMO. These XRPs, implemented in Anglican, can also be re-used.

### 3.1.1 Data-driven inference for DDPMO

In DDPMO, we focus on improving a particular proposal of cluster assignment for a new data point (foreground pixel).

The features $\phi(\rho_k)$ of the environment $\rho_k = (\mathbf{x}_{k-1}, y_{1:k})$, which are the inputs to the neural network, consist of the following:

- Distances to the three nearest clusters in the ascending order, $d_i \in \mathbb{R}, i = 1, \ldots, 3$.
- Colour histograms of a $7 \times 7$ patch surrounding these three clusters in the discretised HSV space, normalised to sum to one, $h_i \in \mathbb{R}^{10}, \sum_{j=1}^{10} h_{ij} = 1$.
- Colour histogram of a $7 \times 7$ patch surrounding the new data point (i.e. pixel) in the discretised HSV space, normalised to sum to one, $c \in \mathbb{R}^{10}, \sum_{i=1}^{10} c_i = 1$.

The outputs of the neural network should be the probabilities of choosing one of existing clusters, or a completely new one. Thanks to the features $\phi(\rho_k)$ that identify the three closest clusters, we use the neural network with only five outputs. These outputs are the probabilities $p_{1:3}$ of choosing the three nearest clusters, the probability $p_4$ for the remaining $K-3$ existing clusters (so that each one has probability $p_4/(K-3)$), and the probability $p_5$ for the entirely new cluster. We directly set the weights of the proposal distribution $\hat{q}(\mathbf{x}_n|\eta)$ to the softmax output of the neural network. If $K < 3$, the prior proposal is used. If $K = 3$, the probability $p_4$ is set to zero (all other probabilities are re-normalised).

The cost function which is used for training of the neural network is the negative log probability given in (1). Noting that the weights are identical, we can use the negative log of the softmax output, and hence we can use neural network packages out of the box.

### 3.1.2 Football

For our experiments, a soccer video dataset was chosen [1]. This choice was made because the video contains many fast-moving, differently coloured and occluding objects. In addition, there exists a human-annotated ground-truth for this dataset. We select two subsequences of frames to form a training dataset and a test dataset. Both datasets mostly consist of moments of intensive play with many players on the field. To measure the performance, we use and report commonly used performance metrics: the sequence frame detection accuracy (SFDA) for object detection and the average tracking accuracy (ATA) for tracking [5].

At first, we run several iterations of SMC inference in Anglican for this model given the input frames from the formed training dataset, with 5000 particles. This allows us to extract inputs and outputs for the neural network, as described in the previous section. Then we train a neural network[1] using these extracted data. Once the neural network is trained, we run inference on the test frame sequences. We measure inference performance with three different types of proposals: the DDPMO prior proposal (i.e. just following the generative model), the data-driven proposal with the probabilities $p_{1:5}$ from the neural network, and the hand-tuned data-driven proposal with fixed probabilities $p_{1:5}$ from empirical analysis. The hand-tuned fixed probabilities $p_{1:5}$ approximate the distribution over the output over the three closest clusters, remaining old clusters, and a new cluster. Finally, we examine how the performance of these methods differ as we change the number of SMC particles.

---

[1] We used a feedforward neural network: one hidden layer with 100 nodes, `tansig` transfer function from the input to the hidden layer, softmax transfer function to the output, and cross-entropy error.

Figures 1 and 2 illustrate the experimental results. For inference with few particles, we get significant improvement in performance using the data-driven proposal. With respect to the particle log-weight, the SMC inference with 10 particles with the data-driven proposal produces results similar to the results from running SMC with thousands of particles under the prior proposal. Thus, using the data-driven proposal, the inference explores the high-probability regions in the posterior space much faster than otherwise.

With respect to performance metrics, for few particles, the performance of SMC with the data-driven proposal is significantly better in comparison to the SMC with the prior proposal with the same number of particles. However, the improvement is less significant, especially in respect to the SFDA metric. In addition, with many particles, SMC with the prior proposal outperforms SMC with the data-driven proposal.

Also, in general, data-driven proposals with the neural network show the same performance as the data-driven proposal with a hand-tuned discriminative model that always returns fixed $p_{1:5}$. However, for the case of SFDA metric performance on the test dataset, the hand-tuned proposal outformed the data-driven proposal with the neural network.

In addition, it is worth noting that even when we attempted to decrease $p^*$ (thus increasing the probability of using the prior proposal), the SFDA metric values for SMC, with the data-driven proposals with 100 particles and more, did not become better for the test dataset and remained very similar to what we see in Figure 2. This might mean that, even though the data-driven proposal allows inference to find high-probability posterior regions much faster and with much less computation effort (as shown in "Log-weight" subfigure in Figure 2), it is not necessarily the case that all performance metrics of interest will be high for samples from those high-probability posterior regions. On the other hand, the last statement is apparent since the generative model is always only a simplification of the real process. Future experiments might be helpful to provide more experimental details on this.

Examples of frames with detected and tracked objects are provided in Figure 3 in the appendix.

## 4 Conclusion and Future Work

This abstract presents an approach to use data-driven proposals for Bayesian non-parametric models in probabilistic programming settings. Our experimental results show that the data-driven proposal significantly improves the inference perfomance. We assume that our proposal might be applied to non-parametric generative models that contain some distance function between clusters and data points (i.e. observations).

The data-driven proposal, which we presented, relies on the feature extractor. The feature extractor maps the current state of the unbounded number of clusters with their sufficient statistics to the input of the neural network. The feature extractor that we implement and use is also the significant part of the data-driven proposal. This is proved by the fact that the neural network performs as well as the fixed hand-tuned discriminative model. This is probably because the spatial factor is important for the model and the dataset, with which we worked. Therefore, there is future work to verify whether for more complex datasets and models data-driven proposals with neural networks provide more benefits.

Our work relates to other work in the field on data-driven proposals. The work on using discriminative proposals for Markov Chain Monte Carlo in parametric generative models include [8] and [4], with applications in computer vision. Recent work with sequential Monte Carlo includes neural adaptive SMC [3], where authors also adapt proposals by descending the inclusive Kullback-Leibler divergence between the proposal and the true posterior distributions on hidden variables given observations. They use recurrent neural networks to train proposal distributions for inference in parametric generative models with fixed dimensionality. Another related recent work is a new probabilistic programming language Picture [6], for which authors propose and describe how to use data-driven proposals for models in vision. They also use neural networks to learn proposals. To get the data to train the neural network, they sample both hidden variables and observations from the generative model unconditionally offline.

In future work, more recent versions of neural networks architectures can be applied to improve results by extracting better features and processing them more efficiently. In particular, one can think of using convolutional neural networks that process the part of the frame centered at the new observing pixel, or even the whole image.

## Acknowledgments

We are very grateful to Willie Neiswanger, Jan Willem van de Meent and Brooks Paige for their help and advice.

## References

[1] T. D'Orazio, M. Leo, N. Mosca, P. Spagnolo, and P. L. Mazzeo. A semi-automatic system for ground truth generation of soccer video sequences. In *Advanced Video and Signal Based Surveillance, 2009. AVSS'09. Sixth IEEE International Conference on*, pages 559–564. IEEE, 2009.

[2] A. Doucet and A. M. Johansen. A tutorial on particle filtering and smoothing: Fifteen years later.

[3] S. Gu, R. E. Turner, and Z. Ghahramani. Neural adaptive sequential monte carlo. *arXiv preprint arXiv:1506.03338*, 2015.

[4] V. Jampani, S. Nowozin, M. Loper, and P. V. Gehler. The informed sampler: A discriminative approach to bayesian inference in generative computer vision models. *Computer Vision and Image Understanding*, 136:32–44, 2015.

[5] R. Kasturi, D. Goldgof, P. Soundararajan, V. Manohar, J. Garofolo, R. Bowers, M. Boonstra, V. Korzhova, and J. Zhang. Framework for performance evaluation of face, text, and vehicle detection and tracking in video: Data, metrics, and protocol. *Pattern Analysis and Machine Intelligence, IEEE Transactions on*, 31(2):319–336, 2009.

[6] T. D. Kulkarni, P. Kohli, J. B. Tenenbaum, and V. K. Mansinghka. Picture: a probabilistic programming language for scene perception. 2015.

[7] W. Neiswanger, F. Wood, and E. P. Xing. The dependent dirichlet process mixture of objects for detection-free tracking and object modeling. 2014.

[8] Z. Tu and S.-C. Zhu. Image segmentation by data-driven markov chain monte carlo. *Pattern Analysis and Machine Intelligence, IEEE Transactions on*, 24(5):657–673, 2002.

[9] F. Wood, J. W. van de Meent, and V. Mansinghka. A new approach to probabilistic programming inference. In *Proceedings of the 17th International conference on Artificial Intelligence and Statistics*, pages 2–46, 2014.

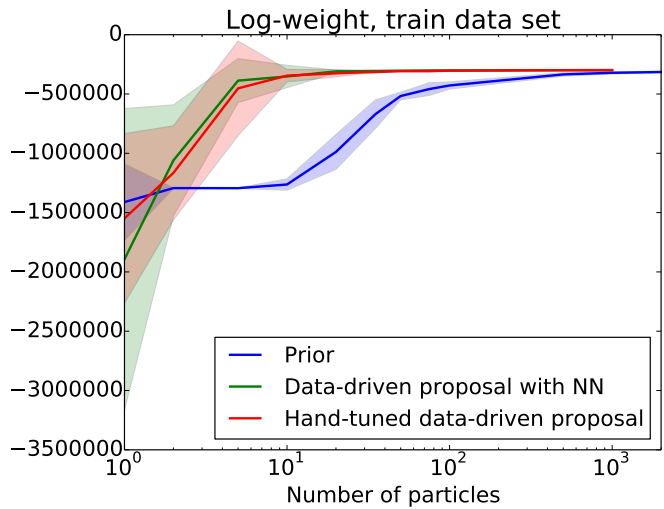

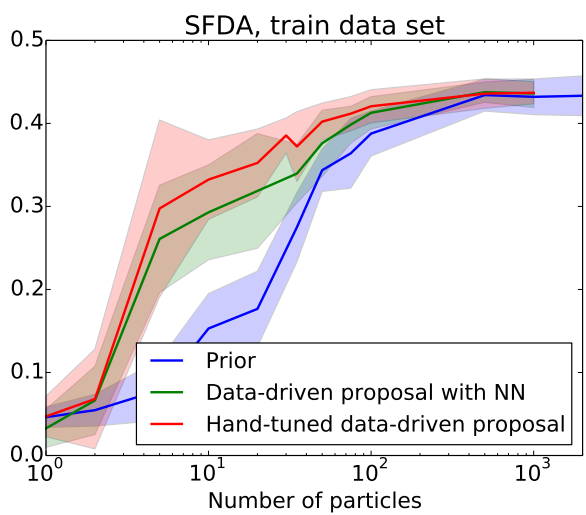

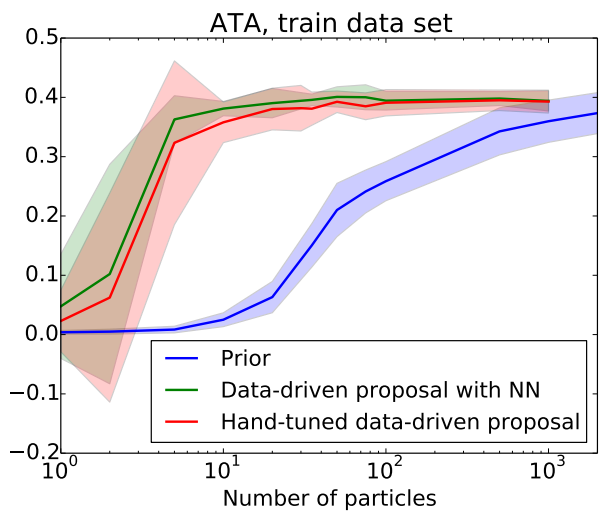

Figure 1: Train dataset. Particle log-weight and perfomance metrics values, namely the sequence frame detection accuracy (SFDA) for object detection and the average tracking accuracy (ATA), for inference results with different proposal types and different number of particles. For the log-weight and both metrics, the higher value is generally better.

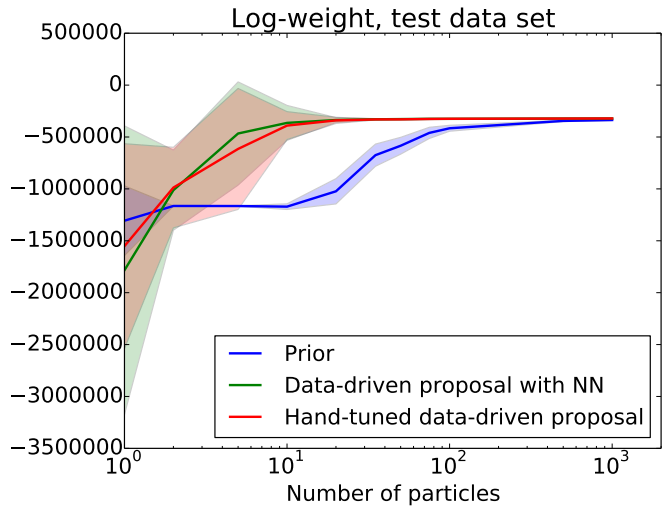

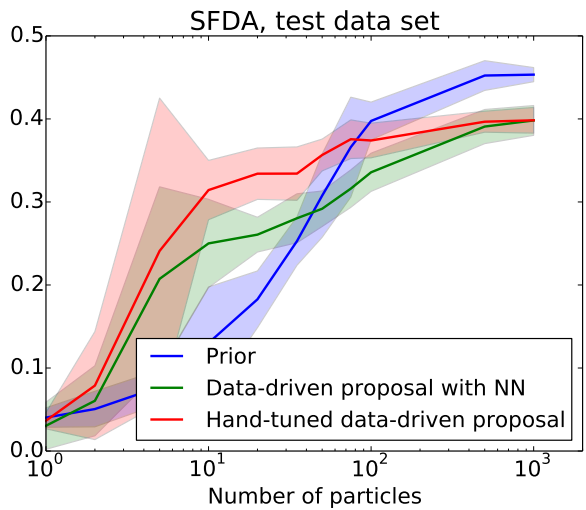

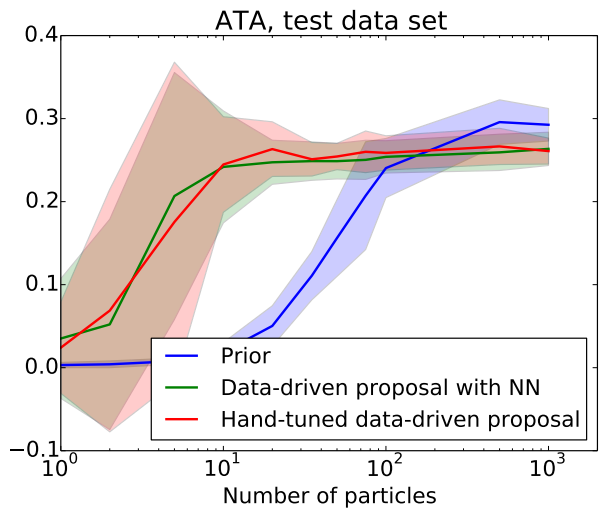

Figure 2: Test dataset. Particle log-weight and perfomance metrics values, namely SFDA and ATA, for inference results with different proposal types and different number of particles. For the log-weight and both metrics, the higher value is generally better.

# Appendix

## 4.1 Object recognition and tracking results with the DDPMO

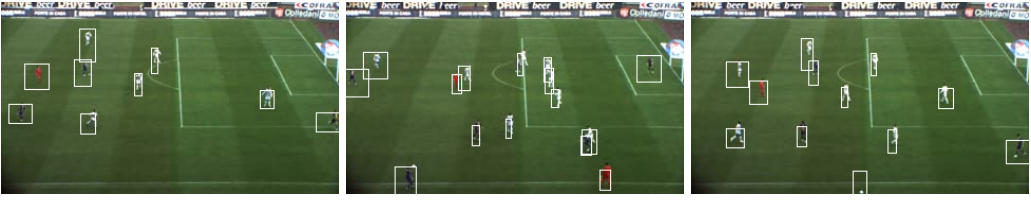

Figure 3: Frames of a soccer video dataset with detected and tracked objects using the DDPMO model in the probabilistic programming system Anglican.

## 4.2 The dependent Dirichlet process mixture of objects (DDPMO) model

The DDPMO models the position and colour $\mathbf{x}_{t,n}$ of a foreground pixel $n$ at a video frame $t$ as an observed variable. This observed variable $\mathbf{x}_{t,n}$ depends on the latent variables of the model such as cluster assignments $c_{t,1:N_t}$ and object parameters $\theta_t^k$ for each cluster $k$. Let $N_t$ be the total number of foreground pixels at a video frame $t$. Then the generative process for each time step $t = 1, \ldots, T$ is as follows:

1. Draw $\{c_{t,1:N_t}, K_{t,N_t}, m_{t,0}^{1:K_{t-1,N_{t-1}}}\} \sim \text{GPU}(\alpha, \rho)$.
2. For $k = 1, \ldots, K_{t,N_t}$:
$$\text{draw } \theta_t^k \sim \begin{cases} \text{T}(\theta_{t-1}^k) & \text{if } k \leq K_{t-1,N_{t-1}} \\ \mathbb{G}_0(\boldsymbol{\eta}_0) & \text{if } k > K_{t-1,N_{t-1}}. \end{cases}$$
3. For $n = 1, \ldots, N_t$: draw $\mathbf{x}_{t,n} \sim \text{F}(\theta_t^{c_{t,n}})$.

Here, $K_{t,n}$ represents the total number of clusters after processing pixel $\mathbf{x}_{t,n}$ and $m_{t,n}^k$ represents the size of cluster $k$ at time after processing $\mathbf{x}_{t,n}$. $\text{F}(\theta_t^k)$ is a generative model that generates the foreground pixels given the object parameter $\theta_t^k$ of cluster $k$. Distributions T and $\mathbb{G}_0$ are transition and prior distributions of the object parameters $\theta_t^k$, which must satisfy a technical condition of the GPU, such that $\int \mathbb{G}_0(\theta_{t-1}^k) \, \text{T}\left(\theta_t^k \middle| \theta_{t-1}^k\right) \mathrm{d}\theta_{t-1}^k = \mathbb{G}_0(\theta_t^k)$. The dependent Dirichlet process prior $\text{GPU}(\alpha, \rho)$ is parametrised by the birth and deletion rates $\alpha$ and $\rho$, which, for each time $t$, governs the evolution of the number of clusters. In this model, the target quantity to infer is

$$p\left(\{c_{t,1:N_t}, K_{t,N_t}, \theta_t^{1:K_{t,N_t}}\}_{t=1:T} \middle| \{\mathbf{x}_{t,n}\}_{t=1:T, n=1:N_t}\right).$$

## 4.3 Generalised Pólya urn

The generative model of the GPU at each time step $t$ is as follows:

1. For $k = 1, \ldots, K_{t-1,N_{t-1}}$:
   (a) Draw $\Delta m_{t-1}^k \sim \text{Binomial}(m_{t-1,N_{t-1}}^k, \rho)$.
   (b) Set $m_{t,0}^k = m_{t-1,N_{t-1}}^k - \Delta m_{t-1}^K$.
2. For $n = 1, \ldots, N_t$
   (a) Draw $c_{t,n} \sim \text{Categorical}\left(\frac{m_{t,n-1}^1}{\alpha + \sum_k m_{t,n-1}^k}, \ldots, \frac{m_{t,n-1}^{K_{t,n-1}}}{\alpha + \sum_k m_{t,n-1}^k}, \frac{\alpha}{\alpha + \sum_k m_{t,n-1}^k}\right)$.
   (b) If $c_{t,n} \leq K_{t,n-1}$, set $m_{t,n}^{c_{t,n}} = m_{t,n-1}^{c_{t,n}} + 1$, $m_{t,n}^{\backslash c_{t,n}} = m_{t,n-1}^{\backslash c_{t,n}}$ and $K_{t,n} = K_{t,n-1}$;[2]
   (c) Otherwise, set $m_{t,n}^{c_{t,n}} = 1, m_{t,n}^{\backslash c_{t,n}} = m_{t,n-1}^{\backslash c_{t,n}}$ and $K_{t,n} = K_{t,n-1} + 1$.

---

[2] $m_{t,n}^{\backslash c_{t,n}}$ is the set $\{m_{t,n}^1, \ldots, m_{t,n}^{K_{t,n}}\} \setminus \{m_{t,n}^{c_{t,n}}\}$.

## 4.4 The DDPMO code in Anglican

```clojure
 1  (ns ddpmo.ddpmo
 2    (:use [anglican emit runtime]
 3           [anglib xrp utils new-dists anglican-utils]
 4           ddpmo.ddpmo-header)
 5    (:require [clojure.core.matrix :as m]
 6              [clojure.core.matrix
 7               :refer [identity-matrix mmul add sub transpose matrix to-nested-vectors]
 8               :rename {identity-matrix eye
 9                        add madd
10                        sub msub
11                        transpose mtranspose}]
12              [clojure.core.matrix.linear :as ml]))
13
14  (with-primitive-procedures
15    [multivariate-t mvn-conjugate-fast dirichlet-multinomial-process
16     DIRICHLET-MULTINOMIAL-PROCESS-STATE-INFO MVN-PROCESS-FAST-STATE-INFO
17     matrix produce-matrix-from-vector to-nested-vectors mtranspose matrix-to-clojure-vector]
18
19    (defquery ddpmo
20      "The Dependent Dirichlet Process Mixture of Objects for Detection-free Tracking"
21      [data Nts proposal-type]
22
23      (let [
24
25            ;;;;;; DDPMO model ;;;;;;
26
27            ;; Hyperparameters for squares/objects/football
28            alpha 0.1 ; for GPU
29            rho 0.32 ; for GPU
30            mu-0 (produce-matrix-from-vector [0 0]) ; for normal-inverse-wishart
31            k-0 0.00370790649926 ; for normal-inverse-wishart
32            nu-0 7336.3104796 ; for normal-inverse-wishart (old prior 60)
33            Lambda-0 (matrix [[193.362493995 0] [0 40.6543682123]])
34            q-0 (vec (repeat 10 10.0)) ; for Dirichlet.
35                                        ; The dimensionality must match number of RGB bins V
36            M 10.1 ; for G0 (eqns (7-8))
37            multinomial-trials 49 ; for eqn (2)... this is m x m where m = 2L + 1
38
39            extract-old-style-theta
40            (fn [theta]
41              (let
42                [mvn-process (retrieve (get theta 'positions))
43                 dirichlet-multinomial-process-instance (retrieve (get theta 'colours))
44                 mu-Sigma (MVN-PROCESS-FAST-STATE-INFO mvn-process)
45                 ps (DIRICHLET-MULTINOMIAL-PROCESS-STATE-INFO
46                      dirichlet-multinomial-process-instance)
47                 theta]
48                {'mu (get mu-Sigma 'mu) 'Sigma (get mu-Sigma 'Sigma)
49                 'trials multinomial-trials 'ps ps}))
50
51            get-N (fn [t] (nth Nts (dec t)))
52
53            ;; Transition distribution
54            T (fn T [prev-theta]
55                (let [previous-mvn-process
56                      (get prev-theta 'positions)
57                      previous-dirichlet-multinomial-process (get prev-theta 'colours)
58
59                      new-mvn-process (XRP (mvn-conjugate-fast mu-0 k-0 nu-0 Lambda-0))
60                      new-dirichlet-multinomial-process
61                      (XRP (dirichlet-multinomial-process q-0 multinomial-trials))
62
63                      ;; Auxiliary transition
```

```
64                      _ (repeatedly M (fn []
65                                        (INCORPORATE new-mvn-process
66                                                     (SAMPLE previous-mvn-process))))
67                      _ (repeatedly M (fn []
68                                        (INCORPORATE
69                                          new-dirichlet-multinomial-process
70                                          (SAMPLE previous-dirichlet-multinomial-process))))
71                      ]
72                  {'positions new-mvn-process 'colours new-dirichlet-multinomial-process}))

74          ;; Base distribution
75          G0 (fn G0 []
76              (let [mvn-process
77                    (XRP (mvn-conjugate-fast mu-0 k-0 nu-0 Lambda-0))
78                    dirichlet-multinomial-process-instance
79                    (XRP (dirichlet-multinomial-process q-0 multinomial-trials))
80                    ]
81                {'positions mvn-process 'colours dirichlet-multinomial-process-instance}))

83          [gpu get-theta] (create-gpu alpha rho G0 T get-N)

85          ;; Helper function
86          ;; Returns parameters for the corresponding table of foreground pixel n at time t
87          get-theta-t-n (mem (fn get-theta-t-n [t n]
88                               (let [customers (gpu t n)
89                                     cs (get customers 'cs)
90                                     k (get cs (dec n))]
91                                 (get-theta t k))))

93          ;;;;;; OBSERVES ;;;;;;
94          observe-lines
95          (fn observe-lines [lines line-id]
96            (if (nil? (first lines))
97              true
98              (let [line (first lines)
99                    pos (get line 'pos)
100                   _ (store "current-pos" (matrix-to-clojure-vector pos))
101                   col (get line 'col)
102                   _ (store "current-col" col)
103                   t (get line 't)
104                   n (get line 'n)
105                   theta (get-theta-t-n t n)
106                   positions-process (get theta 'positions)
107                   colours-process (get theta 'colours)]

109               ; Observing positions.
110               (OBSERVE positions-process pos)

112               ; Observing colours
113               (OBSERVE colours-process col)

115               (if (= n (get-N t))
116                 (let [gpu (gpu t n)
117                       cs (get gpu 'cs)
118                       relevant-clusters (distinct cs)
119                       thetas (map (fn [k]
120                                     (let [theta (get-theta t k)
121                                           theta (extract-old-style-theta theta)
122                                           mu (get theta 'mu)
123                                           Sigma (get theta 'Sigma)
124                                           ps (get theta 'ps)]
125                                       {'k k 'mu mu 'Sigma Sigma 'ps ps}))
126                                   relevant-clusters)
127                       res {'t t 'n n 'gpu gpu 'thetas thetas}]
128                   (predict res)))
```

```
129                        (observe-lines (rest lines) (inc line-id)))))]
130
131          (observe-lines data 0))))
132
133  (defn -main [data-set-name number-of-particles num-particles-to-output
134               proposal-type & ignore-following-args]
135    (let [number-of-particles (parse-int number-of-particles)
136          num-particles-to-output (parse-int num-particles-to-output)
137          proposal-type (str proposal-type)
138          _ (case proposal-type "prior" :okay "handtuned" :okay "nn" :okay)
139          [data Nts] (load-DDPMO-data data-set-name)
140          query-results (doquery :smc ddpmo [data Nts proposal-type]
141                                 :number-of-particles number-of-particles)
142          results
143          (doall
144            (map
145              (fn [particle-output particle-id]
146                (doall
147                  (map
148                    (fn [x]
149                      (println (str particle-id "," (first x) "," (second x) ",0.0")))
150                    (get particle-output :anglican.state/predicts))))
151              (take num-particles-to-output query-results)
152              (range num-particles-to-output)))]
153      results))
```

### 4.5 The GPU code in Anglican

```
1   ;;;;;; GPU definition ;;;;;;
2
3   ; Creates an instance of a GPU process.
4   ; Takes:
5   ; * GPU's alpha and rho.
6   ; * Base distribution G0.
7   ; * Transition distribution T.
8   ; * function get-N which returns the number of points at each time.
9   ; Returns: [gpu get-theta]
10  (defm create-gpu [alpha rho G0 T get-N]
11    (let
12      [;; Given vector of table sizes ms = [m1 m2 ...], returns a new vector of table
13       ;; sizes by removing customers from tables with probability rho
14       remove-customers
15       (fn [ms]
16         (vec (map (fn [m]
17                     (if (= m 0) 0 (- m (SAMPLE (binomial m rho)))))
18                   ms)))
19
20       ;; Returns {'cs (vector of n cluster ids) 'K (number of unique clusters at n)
21       ;;          'ms (vector of cluster sizes at n)}
22       ;; after processing foreground pixel n at time t
23       ;; n goes from 1
24       ;; c_i goes from 0
25       ;; K = max(c_i) + 1
26       ;; t goes from 1
27       gpu (mem
28             (fn gpu [t n]
29               (if (= n 0)
30                 ;; Initialise
31                 (if (= t 1)
32                   {'cs '[] 'K 0 'ms '[]}
33                   (let [prev-t-gpu (gpu (dec t) (get-N (dec t)))
34                         prev-K (get prev-t-gpu 'K)
35                         prev-ms (get prev-t-gpu 'ms)]
36                     {'cs '[] 'K prev-K 'ms (remove-customers prev-ms)}))
```

```
37
38                ;; Get from step (n - 1)
39                (let [prev-n-gpu (gpu t (dec n))
40                      cs (get prev-n-gpu 'cs)
41                      K (get prev-n-gpu 'K)
42                      ms (get prev-n-gpu 'ms)
43                      w (conj ms alpha)
44                      c (SAMPLE (discrete w))
45                      new-cs (conj cs c)
46                      new-K (max K (inc c))
47                      new-ms (assoc ms c (inc (get ms c 0)))]
48                  {'cs new-cs 'K new-K 'ms new-ms})))
49
50        ;; Returns parameters for table k at time t using either
51        ;; transition distribution T or base distribution GO
52        get-theta (mem (fn get-theta [t k]
53                        (if (= t 1)
54                          (GO)
55                          (let [prev-customers (gpu (dec t) (get-N (dec t)))
56                                prev-K (get prev-customers 'K)
57                                initial-ms (get (gpu t 0) 'ms)]
58                            (if (> k (dec prev-K))
59                              (GO)
60                              (if (= (nth initial-ms k) 0)
61                                nil
62                                (T (get-theta (dec t) k))))))))]
63     [gpu get-theta]))
```

### 4.6 Clojure code for the data-driven proposal

```
1  (def NUMBER-OF-NEAREST-CLUSTERS 3)
2
3  (def sort-thetas
4    (fn [thetas]
5      (let
6        [my-comparer
7         (fn [el1 el2]
8           (< (nth el1 2) (nth el2 2)))]
9        (sort my-comparer thetas))))
10
11 (def distance
12   (fn [[x1 y1] [x2 y2]]
13     "Returns Euclidean distance between two 2D points."
14     ; Important! Here x is really y, and vice versa.
15     ; This is because in the MATLAB code the first coordinate is y.
16     (pow (+ (pow (- x1 x2) 2.0) (pow (- y1 y2) 2.0)) 0.5)))
```

### 4.7 Anglican code (within the DDPMO model) for the data-driven proposal

```
1  get-thetas
2  (fn [t n]
3    "Returns thetas for active clusters (ms[i] > 0)
4    at data point (t, n). This function should be
5    called only when we already processed that data point."
6    (let
7      [
8       gpu-state (gpu t n)
9       ms (get gpu-state 'ms)
10      get-theta (fn [t k] (if (> (nth ms k) 0) (get-theta t k) nil))
11      thetas (map (fn [k] (list k (get-theta t k))) (range (count ms)))
12      thetas (filter (fn [el] (not (nil? (second el)))) thetas)
13      ]
14     thetas))
15
```

```
16   get-mean-coords
17   (fn [theta]
18     "Extracts mean from the theta as Clojure vector."
19     (let
20       [coords (matrix-to-clojure-vector
21                (get (MVN-PROCESS-FAST-STATE-INFO
22                      (retrieve (get theta 'positions))) 'mu))]
23       coords))
24
25   get-nearest-thetas
26   (fn [t n [x y]]
27     "Gets an ordered list of theta which are the nearest to the point [x y]
28     based on the state at the previous data point (t, n - 1)."
29     (if (and (= t 1) (= n 1))
30         nil
31         (let
32           [[t n]
33            (if (= n 1)
34                [(- t 1) (get-N (- t 1))]
35                [t (- n 1)])]
36           (let
37             [thetas (get-thetas t n)
38              thetas (map (fn [[k theta]]
39                            (list k theta (distance [x y]
40                                                    (get-mean-coords theta)))) thetas)
41              thetas (sort-thetas thetas)
42              thetas (take NUMBER-OF-NEAREST-CLUSTERS thetas)]
43             (if (< (count thetas) NUMBER-OF-NEAREST-CLUSTERS)
44                 nil
45                 thetas)))))
46
47   ;; Do the trick to allow mutual recursion.
48   _ (store "get-nearest-thetas" get-nearest-thetas)
```

### 4.8 Code for the GPU, to get data for the proposal for train datasets

```
1    NEAREST-THETAS ((retrieve "get-nearest-thetas") t n (retrieve "current-pos"))
2    for-proposal
3      (map
4       (fn [the-list]
5         (let
6           [theta-id (nth the-list 0)
7            theta (nth the-list 1)
8            distance-to-the-center (nth the-list 2)]
9           (list
10           theta-id
11           (DIRICHLET-MULTINOMIAL-PROCESS-STATE-INFO (retrieve (get theta 'colours)))
12           distance-to-the-center)))
13      NEAREST-THETAS)
14   nn-input
15     (concat
16     (apply concat (doall (map (fn [data]
17                                 (concat (nth data 1) (list (nth data 2)))) for-proposal)))
18     (doall (map (fn [x] (/ x 49.0)) (retrieve "current-col"))))
19   c (if (or (not (= (count nn-input) 43)) (= proposal-type "prior"))
20       (SAMPLE (discrete w))
21       (let
22         [dist (sample-cluster-id nn-input w 0.8
23                                  (map first NEAREST-THETAS) (= proposal-type "handtuned"))
24          [my-sample log-likelihood] (sample dist)]
25         (add-log-weight log-likelihood)
26         my-sample))
```

## 4.9 Code for the GPU, to use the proposal for test datasets

```
NEAREST-THETAS ((retrieve "get-nearest-thetas") t n (retrieve "current-pos"))
for-proposal
  (map
    (fn [the-list]
      (let
        [theta-id (nth the-list 0)
         theta (nth the-list 1)
         distance-to-the-center (nth the-list 2)]
        (list
          theta-id
          (DIRICHLET-MULTINOMIAL-PROCESS-STATE-INFO (retrieve (get theta 'colours)))
          distance-to-the-center)))
    NEAREST-THETAS)
_ (predict (list for-proposal c (retrieve "current-col") (count w)))
```

**Revision as of the 10th of April 2016:**

It has been found that there was a bug in the implementation of the multivariate normal XRP. The experiments have been re-run accordingly. The section on experiment result has been updated.